\documentclass[conference]{IEEEtran}
\IEEEoverridecommandlockouts
\usepackage{cite}
\usepackage{amsmath,amssymb,amsfonts}
\usepackage{algorithmic}
\usepackage{graphicx}
\usepackage{textcomp}
\usepackage{xcolor}
\usepackage{booktabs}
\usepackage{color}
\usepackage{hyperref}
\hypersetup{
colorlinks=true,
linkcolor=black
}

\def\BibTeX{{\rm B\kern-.05em{\sc i\kern-.025em b}\kern-.08em
    T\kern-.1667em\lower.7ex\hbox{E}\kern-.125emX}}
\begin{document}

\title{FF-PNet: A Pyramid Network Based on Feature and Field for Brain Image Registration\\
}

\author{
Ying Zhang\textsuperscript{1}, Shuai Guo\textsuperscript{1}, Chenxi Sun\textsuperscript{1},
Yuchen Zhu\textsuperscript{1},
Jinhai Xiang\textsuperscript{1,2,3 
 }$^\dagger$\thanks{$^\dagger$Corresponding author},  \\ 
\textsuperscript{1}\textit{College of Informatics, Huazhong Agricultural University, Wuhan 430070, China}\\
\textsuperscript{2}\textit{Agricultural Bioinformatics Key Laboratory of Hubei Province,}\\
\textsuperscript{3}\textit{Key Laboratory of Smart Farming for Agricultural Animals, Ministry of Agriculture,}\\
\textit{Huazhong Agricultural University, Wuhan 430070, China} \\
\texttt{\{zy,jimmy\_xiang\}@mail.hzau.edu.cn} \\
\texttt{\{2023317120031,linkscx,gjarvey\}@webmail.hzau.edu.cn}
}

\maketitle

\begin{abstract}
In recent years, deformable medical image registration techniques have made significant progress. However, existing models still lack efficiency in parallel extraction of coarse and fine-grained features. To address this, we construct a new pyramid registration network based on feature and deformation field (FF-PNet). For coarse-grained feature extraction, we design a Residual Feature Fusion Module (RFFM), for fine-grained image deformation, we propose a Residual Deformation Field Fusion Module (RDFFM). Through the parallel operation of these two modules, the model can effectively handle complex image deformations. It is worth emphasizing that the encoding stage of FF-PNet only employs traditional convolutional neural networks without any attention mechanisms or multilayer perceptrons, yet it still achieves remarkable improvements in registration accuracy, fully demonstrating the superior feature decoding capabilities of RFFM and RDFFM. We conducted extensive experiments on the LPBA and OASIS datasets. The results show our network consistently outperforms popular methods in metrics like the Dice Similarity Coefficient. \textit{If the paper is accepted, the experimental code will be published on GitHub.}
\end{abstract}


\section{Introduction}
In the field of medical image processing, image registration has long been an active research area. Image registration techniques are generally classified into rigid/affine and non-rigid/deformable categories. Rigid registration is suitable for aligning global rigid body motions, while non-rigid registration addresses the fine-grained alignment of local complex deformations. Deformable image registration, focusing on establishing nonlinear dense correspondences between image pairs, plays a critical role in surgical guidance\cite{matinfar2023sonification}, disease diagnosis\cite{sotiras2013deformable}, and postoperative recovery\cite{siddiqui2015comparison}.
\begin{figure}[hbtp]
\centerline{\includegraphics[width=0.5\textwidth]{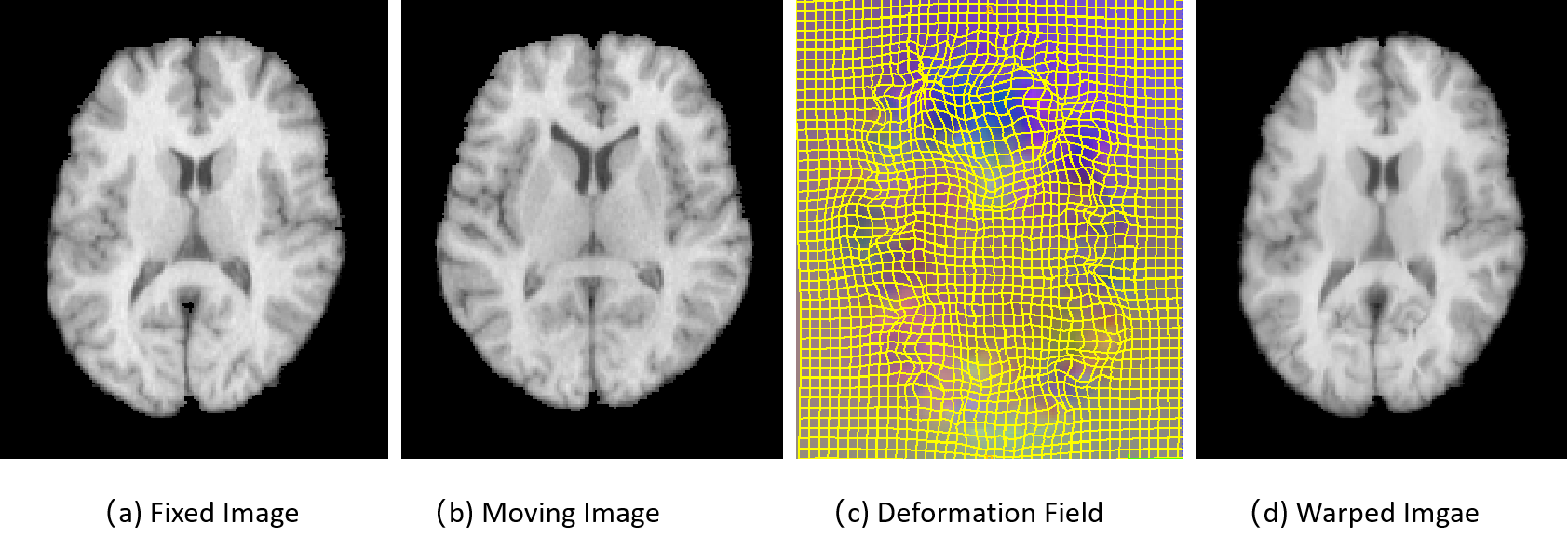}}
\caption{One example to show the deformable image registration. 
a non-rigid deformation field (c) is estimated to warp the moving image (b) to match with the fixed image (a). (d) shows the warped moving image.}
\label{introduction}
\end{figure}

As shown in Figure \ref{introduction}, the core of deformable image registration lies in learning the spatial correspondences between image pairs to generate a non-rigid deformation field, which precisely maps all corresponding anatomical structures from the two images into a unified spatial coordinate system. This process essentially solves for the optimal spatial correspondence between a fixed image and a moving image, formulated as an iterative optimization problem. Traditional registration methods treat deformable registration as an iterative optimization of the deformation field\cite{klein2009evaluation}\cite{rueckert2006diffeomorphic}, but when the target image exhibits drastic anatomical deformations, registration efficiency and computational cost increase dramatically.

In recent years, deep learning-based image registration methods have been widely proposed and applied. These approaches directly infer the deformation field between image pairs through convolutional neural networks (CNNs), significantly enhancing registration accuracy\cite{fu2020deep}\cite{balakrishnan2018unsupervised}. To address the time-consuming nature of traditional iterative optimization methods, the U-Net\cite{ronneberger2015u} architecture has emerged as a key solution. Its symmetric encoder-decoder structure offers unique advantages in capturing coarse-grained long-range dependencies, leading to the widespread adoption of its variants in the registration field\cite{hoopes2021hypermorph}\cite{zhang2024dual}\cite{zhang2023ccac}. The VoxelMorph\cite{balakrishnan2019voxelmorph} is a prominent example, mapping spatial correspondences between image pairs to a deformation field directly via CNNs. However, constrained by the inherent receptive field limitations of CNNs, researchers have integrated vision transformer (ViT)\cite{dosovitskiy2020image} into the U-Net architecture. Leveraging the self-attention mechanism to expand the receptive field, these models efficiently model large-scale deformations between images. TransMorph\cite{chen2022transmorph}, a representative application of transformer\cite{vaswani2017attention} in image registration, endows the model with the ability to capture long-range dependencies through self-attention, addressing the limitations of traditional CNNs. Another category of solutions employs feature pyramid strategies\cite{kang2022dual}\cite{liu2022coordinate}\cite{wang2023modet}\cite{ma2023pivit}, achieving effective capture of fine-grained features through bottom-up multi-stage deformation field prediction in the decoder stage. Although both registration architectures have significantly improved accuracy, the potential of current networks in matching complex anatomical structures remains underutilized.

In this paper, we introduce FF-PNet, an unsupervised pyramid registration network that integrates the strengths of the U-Net architecture and pyramid-based networks to perform registration via feature-level representations and deformation-field modeling. By expanding the receptive field, the model simultaneously enhances its capability to capture local deformations. The decoding process incorporates two novel fusion modules: the Residual Feature Fusion Module (RFFM), which operates on contextual features to boost the model’s decoding efficiency for core semantic information, and the Residual Deformation Field Fusion Module (RDFFM), which acts on multi-scale deformation fields to sharpen edge feature perception. Leveraging these dual-stream fusion mechanisms, FF-PNet exploits the correlation between images and registration stages to achieve coarse-to-fine precise alignment. The main contributions of our work can be summarized as follows:
\begin{itemize}
\item[$\bullet$] We propose a Residual Feature Fusion Module (RFFM), which significantly strengthens the multi-scope dependencies among contextual features by distorting the fused features, thereby improving the feature representation capability.
\item[$\bullet$] We present a Residual Deformation Field Fusion Module (RDFFM), which utilizes cross-layer semantic propagation to capture subtle local deformation displacements, improving deformation field accuracy.
\item[$\bullet$] We design FF-PNet, a three-layer pyramid architecture based on a dual-stream decoder, which adopts a pure convolutional pyramid design to achieve precise image-pair alignment without complex components, demonstrating the efficiency of the proposed fusion modules. Notably, these modules exhibit broad applicability and can be readily integrated into any pyramid-based registration framework, highlighting their generalizable value in medical image registration and beyond.
\end{itemize}

\section{related work}
\subsection{U-Net-based Registration Method}
The U-Net\cite{ronneberger2015u} network is highly esteemed for its symmetric encoder-decoder architecture, delivering superior performance in image segmentation and registration. Its skip connections enable direct linking of feature maps from the encoder to the decoder, merging detailed encoder information during decoding and supplying the decoder with rich contextual data. VoxelMorph\cite{balakrishnan2019voxelmorph}, a popular CNN-based registration network, mirrors this architecture with its hierarchical encoder-decoder setup. Dalca et al. advanced VoxelMorph by introducing a diffeomorphic version\cite{dalca2018andm}, employing scaling and squaring to approximate static velocity field integration and ensure diffeomorphic deformation fields. Building on this, subsequent studies maintained the encoder-decoder structure and integrated Transformer\cite{vaswani2017attention} technology. The self-attention mechanism in Transformers overcame CNN limitations, especially in capturing long-range information in deformation fields. Vision Transformer (ViT)\cite{dosovitskiy2020image} pioneered the use of Transformers in image encoding, and Swin Transformer\cite{liu2021swin} enhanced ViT with windowed attention and a hierarchical structure, boosting computational efficiency for high-resolution images. For example, Zhu et al. presented Swin-VoxelMorph\cite{zhu2022swin}, a purely Transformer-based registration network combining Swin Transformer and VoxelMorph benefits. Chen et al. introduced TransMorph\cite{chen2022transmorph}, a hybrid CNN-Transformer Unet registration network using multiple Swin Transformer blocks in encoding. All these methods build upon the U-Net architecture and share the common feature of directly estimating the registration field.
\subsection{Pyramid-based Registration Method}
In the field of deformable image registration, apart from the common Unet-style direct registration architectures, pyramid registration architectures have also garnered significant attention, with their core lying in performing registration steps in a coarse-to-fine manner. The Recursive Cascade Network (RCN)\cite{zhao2019recursive}, as one of the early medical image networks exploring continuous deformation registration, employs VoxelMorph as a subnetwork and leverages a recursive mechanism to achieve progressive registration effects. DualPRNet++\cite{kang2022dual} inputs two images into a shared-parameter U-Net structure, advancing registration layer-by-layer through a Pyramid Registration (PR) module. By using sequential deformation operations in a coarse-to-fine order to continuously optimize the deformation field, it demonstrates strong capabilities in handling large-deformation scenarios. PIVIT\cite{ma2023pivit} constructs a pyramid-based iterative registration framework and innovatively introduces Swin Transformer at the lowest scale of the decoding process. The Recursive Decomposition Network (RDN)\cite{lv2022joint} ingeniously integrates a recursive mechanism into the pyramid architecture. By decomposing large deformation fields into affine matrices and multiple smaller deformable fields, it effectively overcomes the bottleneck in traditional pyramid architectures where the number of deformation fields is limited by resolution. Im2grid\cite{liu2022coordinate} combines neighborhood attention mechanisms to progressively predict fine-grained deformation fields at each hierarchy. ModeT\cite{wang2023modet}, taking a different approach, fully exploits the potential of Transformer structures in deformation estimation by converting multi-head neighborhood attention relationships into multi-coordinate relationships, thus enabling more accurate modeling of correspondences between images and providing new ideas and methods for deformable image registration. However, pyramid-based registration methods still have substantial room for improvement in terms of accuracy, particularly when dealing with extreme large deformations or locally complex structures, where their registration precision may be constrained.

\begin{figure*}[htbp]
\begin{center}
\includegraphics[width=1\textwidth]{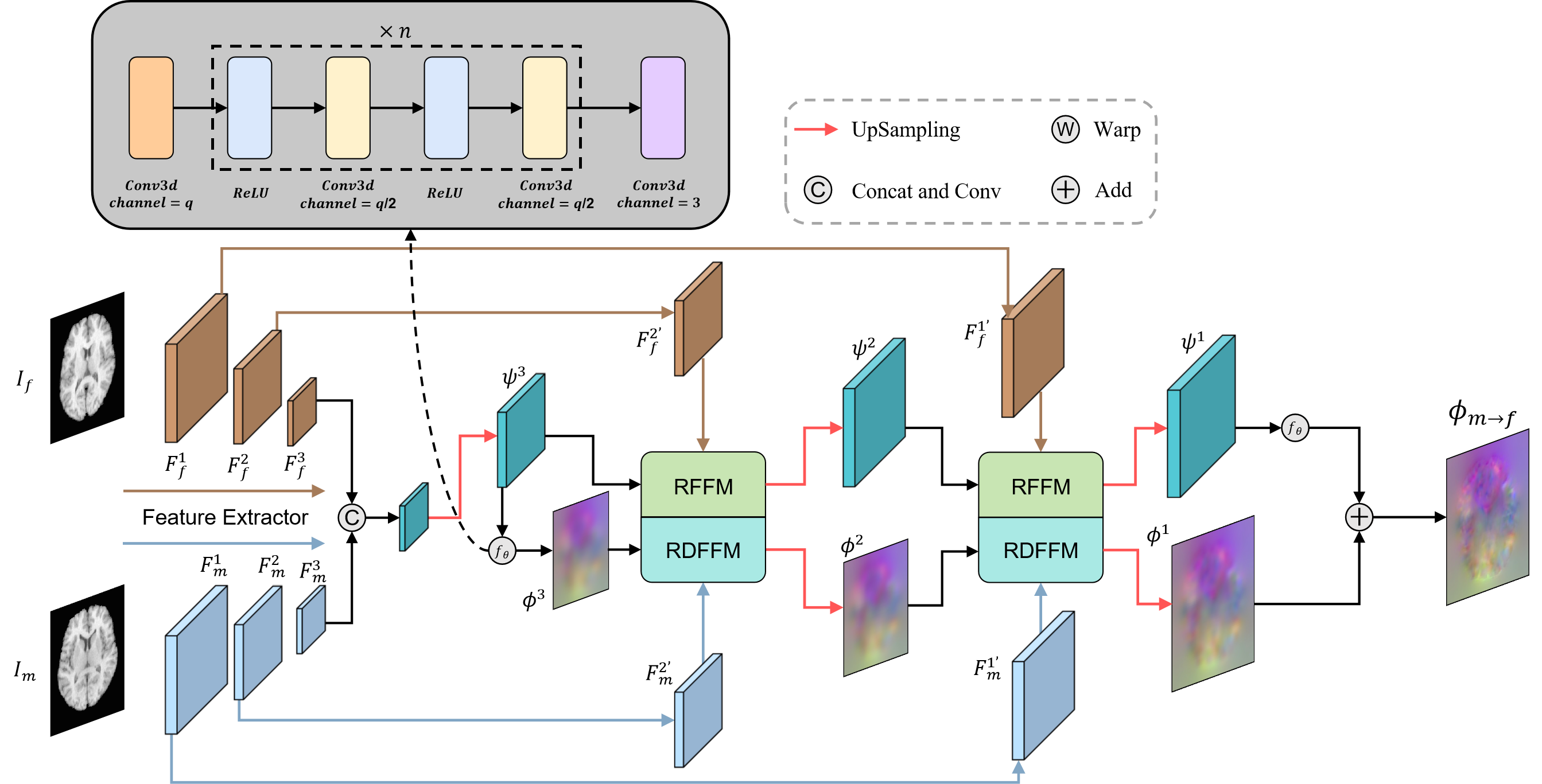}
\end{center}
\caption{The overall architecture of our FF-PNet. It is a three-layer pyramid network composed of a hierarchical feature extraction encoder based on CNN and a coarse-to-fine registration decoder based on two fusion modules (including RFFM and RDFFM).}
\label{fg:method}
\end{figure*}

\section{methods}
We calculate the similarity between the moving image, which has undergone non-rigid deformation by converting the output features of different encoder layers into deformation fields, and the fixed image. Experimental results show that when the network adopts a four-layer architecture, the image distorted by the deformation field of the penultimate layer has the highest similarity to the fixed image compared with the last layer. This indicates that appropriate layer selection is more conducive to capturing key spatial information within the image than blindly increasing the number of layers. The design of the number of layers in a deep learning model needs to balance between complexity and performance to achieve efficient image registration. 
\subsection{The Overview of FF-PNet}\label{AA}
The FF-PNet network architecture depicted in Figure \ref{fg:method} is composed of three-layer feature extraction encoders, a Residual Feature Fusion Module (RFFM), and a Residual Deformation Field Fusion Module (RDFFM). This network employs a weight-sharing feature extractor to construct multi-scale feature pyramids \(\{F_f^i\}\) and \(\{F_m^i\}\) for the fixed image \(I_f\) and the moving image \(I_m\), respectively, where the scale index \(i \in \{1, 2, 3\}\). The feature extractor comprises three cascaded convolutional modules, with adjacent modules connected by a \(2 \times 2 \times 2\) max-pooling layer, and each module consists of a 3D convolutional layer and a Leaky ReLU activation function with a parameter of \(0.1\). To achieve multi-scale feature fusion, the original features \(\{F_f^i\}\) and \(\{F_m^i\}\) are fed into a two-layer decoder. Each layer of the decoder is composed of an up-sampling layer and two consecutive 3D convolutional layers, outputting decoded features \(\{F_f^{i'}\}\) and \(\{F_m^{i'}\}\), where \(i \in \{1, 2\}\), and these decoded features are conveyed to the fusion module for processing. Specifically, at the coarsest scale (\(i = 3\)), FF-PNet does not utilize the fusion module for the features \(F_f^3\) and \(F_m^3\), only performing convolutional operations and up-sampling processing in the channel dimension. This process can be formulated as:
\begin{equation}
\psi^3 = Up(CnC(F_f^3,F_m^3))
\end{equation}

At subsequent scale levels, the fused feature \(\psi^3\) and the decoded features \(\{F_f^{2'}\}\) and \(\{F_m^{2'}\}\) are fed into RFFM and RDFFM. Specifically, the RFFM is dedicated to exploring the principal correspondence relationship in the spatial domain between the fixed image and the moving image, while the RDFFM focuses on extracting the spatial detail information between image pairs. By virtue of the parallel-operation mechanism of the two modules, FF-PNet is enabled to achieve coarse-to-fine registration performance. The structure of \( f_\theta \) is shown in Figure \ref{fg:method}, which is composed of double convolutional blocks stacked for \( n \) layers, used for channel indentation and same-channel feature extraction respectively. At different feature scales, \( f_\theta \) is used to process the fixed image feature \( F_f \) and the moving image \( F_m \), or the fixed image feature \( F_f \) and the warped image feature \( F_w \) to generate a registration field of the same scale.
\begin{equation}
\phi = f_\theta (CnC (F_f , F_m )) \text{ or } f_\theta (CnC (F_f , F_w )) 
\end{equation}

Taking the second layer as an example, the RFFM takes \(\psi^3\) as the main feature flow, and uses \(F_f^{2'}\) and \(F_m^{2'}\) as complementary feature flows to extract the residual feature flow. Eventually, cross-fusion of the three flows is realized to acquire the multi-scope dependencies among features. Notably, \(\psi^3\) is not directly input into the deformation field fusion module. Instead, it is first processed by \(f_\theta\) to generate \(\phi^3\), which serves as the main body of the deformation field flow. Then, \(F_f^{2'}\) and \(F_m^{2'}\) extract the residual field and integrate it into \(\phi^3\). The final registration field is directly formed by the superposition of the registration fields output by the two modules. This design further enhances the model's ability to integrate different levels of information in image registration.

\subsection{Residual Feature Fusion Module}
Currently, most U-Net-based image registration methods adopt complex skip connection configurations or integrate multi-layer perceptrons (MLPs) and attention mechanisms into decoders. While these approaches effectively enhance registration accuracy, they inevitably lead to substantial increases in computational complexity and model parameters, which pose challenges for the practical deployment of image registration techniques. To address this issue, this study introduces an innovative strategy: by simulating deformation operations on decoded features, we extract residual and warped features simultaneously. These features, together with the decoded features, are fed into a cross-fusion module constructed using simple convolutional blocks, which is designed to capture spatial correlations between multi-source features. In summary, the proposed residual feature fusion module, applied during the feature decoding process, achieves efficient feature extraction by significantly reducing both parameter count and computational complexity while maintaining robust performance. 

RFFM contains two core modules, namely the residual feature extraction module and the cross-fusion module. Taking the second layer as an example, RFFM first processes the fused feature \( \psi^3 \) with the function \( f_\theta \) to generate a deformation field, and then performs a warping operation on the moving image feature \( F'^2_m \) to obtain the warped image feature \( F'^2_w \). Subsequently, \( F'^2_w \) and the fixed image feature \( F'^2_f \) are processed by the \( CnC \) module to obtain the fused feature \( \psi^{'} \) of this layer. Then, this fused feature \( \psi^{'} \) uses the \( f_\theta \) function again to generate a deformation field, warping \( \psi^3 \) to obtain the residual feature \( \psi_{\_} \). Finally, the fused feature \( \psi^{'} \), the residual feature \( \psi_{\_} \), and the fixed image feature \( F'^2_f \) are jointly input into the cross-fusion module (CFM), and after processing, the output feature \( \psi^2 \) of this layer is obtained. This process can be represented by the following formula: 
\begin{equation}
    \begin{cases}
    F_w^{2'} = Warp(F_m^{2'},f_\theta(\psi^3))\\
    \psi^{'} = CnC(F_f^{2'},F_w^{2'})\\
    \psi_{\_} = Warp(\psi^3,f_\theta(\psi^{'})\\
    \psi^{2} = CFM(\psi^{'},\psi_{\_},F_f^{2'})\\
    \end{cases}
\end{equation}

\begin{figure}[hbtp]
\centerline{\includegraphics[width=0.5\textwidth]{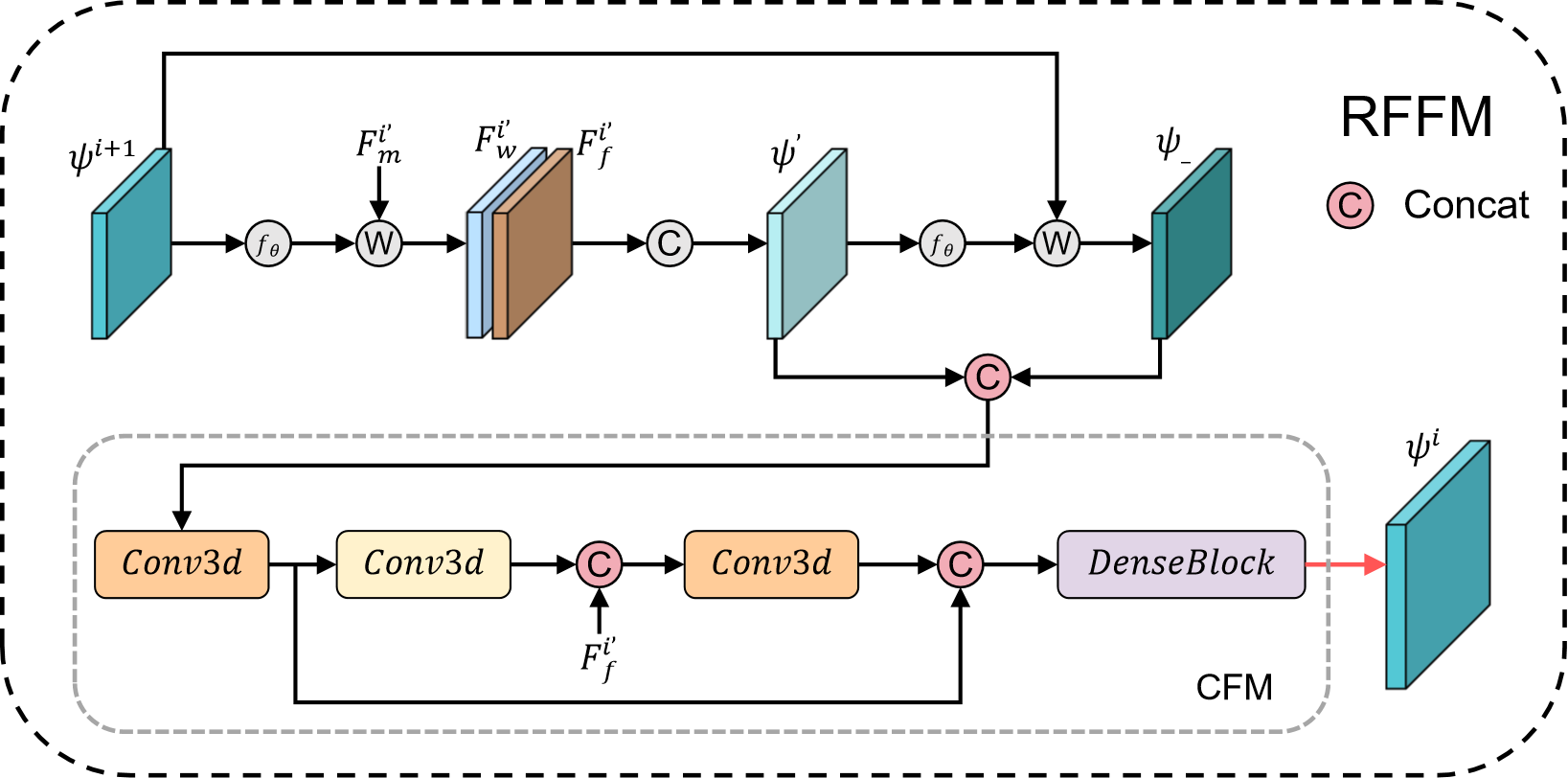}}
\caption{This figure illustrates the detailed implementation of RFFM. The first row in the figure shows the residual feature extraction operations, and the second row describes the composition of the cross-fusion module.}
\label{RFFM}
\end{figure}

Through the CFM, deep cross-feature extraction is performed on the fused feature $\psi^{'}$, residual feature $\psi_{\_}$, and fixed image feature $F_f^{2'}$, enhancing the model’s understanding ability of spatial correspondence and improving the model’s adaptability to complex deformation scenarios. The specific process of the cross-fusion module is as shown in Figure \ref{RFFM}. 

\subsection{Residual Deformation Field Fusion Module}
In the recently conducted pyramidal registration investigations, the necessity of multi-level registration fields has been fully validated. To endow the network with the capability of detail deformation handling, except for the coarsest scale, we emulated the registration process for each layer and named this emulated process as RDFFM. Specifically, RDFFM utilizes \( f_\theta \) to generate the deformation field at the current scale, applies this deformation field to the moving image, and then, with the aid of the resulting warped image and the fixed image, predicts a new deformation field through the same generalized registration function. Notably, FF-PNet performs only straightforward channel convolution and upsampling operations at the coarsest scale. Its output result, after being processed by \( f_\theta \), yields \( \phi^3 \), which serves as the input for RDFFM. The deformation field fusion process of it in the second layer is shown as the following equation: 
\begin{equation}
    \begin{cases}
    \phi^{3} = f_\theta(\psi^3)\\
    F_w^{2'} = Warp(F_m^{2'},\phi^3)\\
    \phi^{'} = f_\theta(CnC(F_f^{2'},F_w^{2'}))\\
    \phi^{2} = Warp(\phi^3,\phi^{'})+\phi^{'}\\
    \end{cases}
\end{equation}

where \(F_f^{2'}\) and \(F_m^{2'}\) denote the decoded features processed by skip connections, \( \phi^{'}\) represents the intermediate-time deformation field, \(CnC\) signifies the concatenation and convolution operations, and \(Warp\) stands for the warping operation. Through this fusion step, the obtained deformation field starts to acquire the capability of identifying local image deformations. To achieve high-precision registration, we repeat this process. After multiple iterations of warping and fusion, the output \(\phi^2\) serves as the input for the subsequent layer.
\begin{figure}[hbtp]
\centerline{\includegraphics[width=0.5\textwidth]{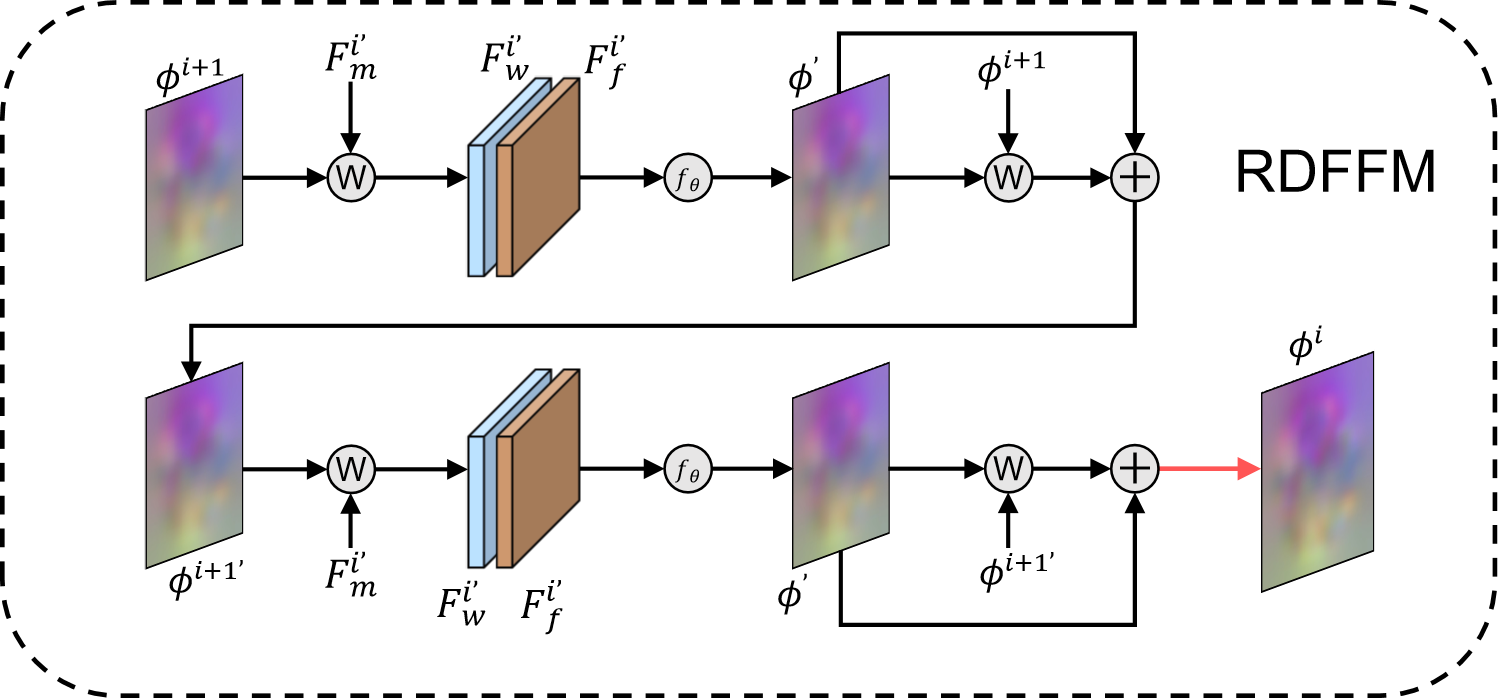}}
\caption{This figure details the process of RDFFM. It performs warping and fusion operations on the input deformation field, and after two iterations, outputs the deformation field via upsampling.}
\label{RDFFM}
\end{figure}

\begin{figure*}[htbp]
\begin{center}
\centerline{\includegraphics[width=1\linewidth]{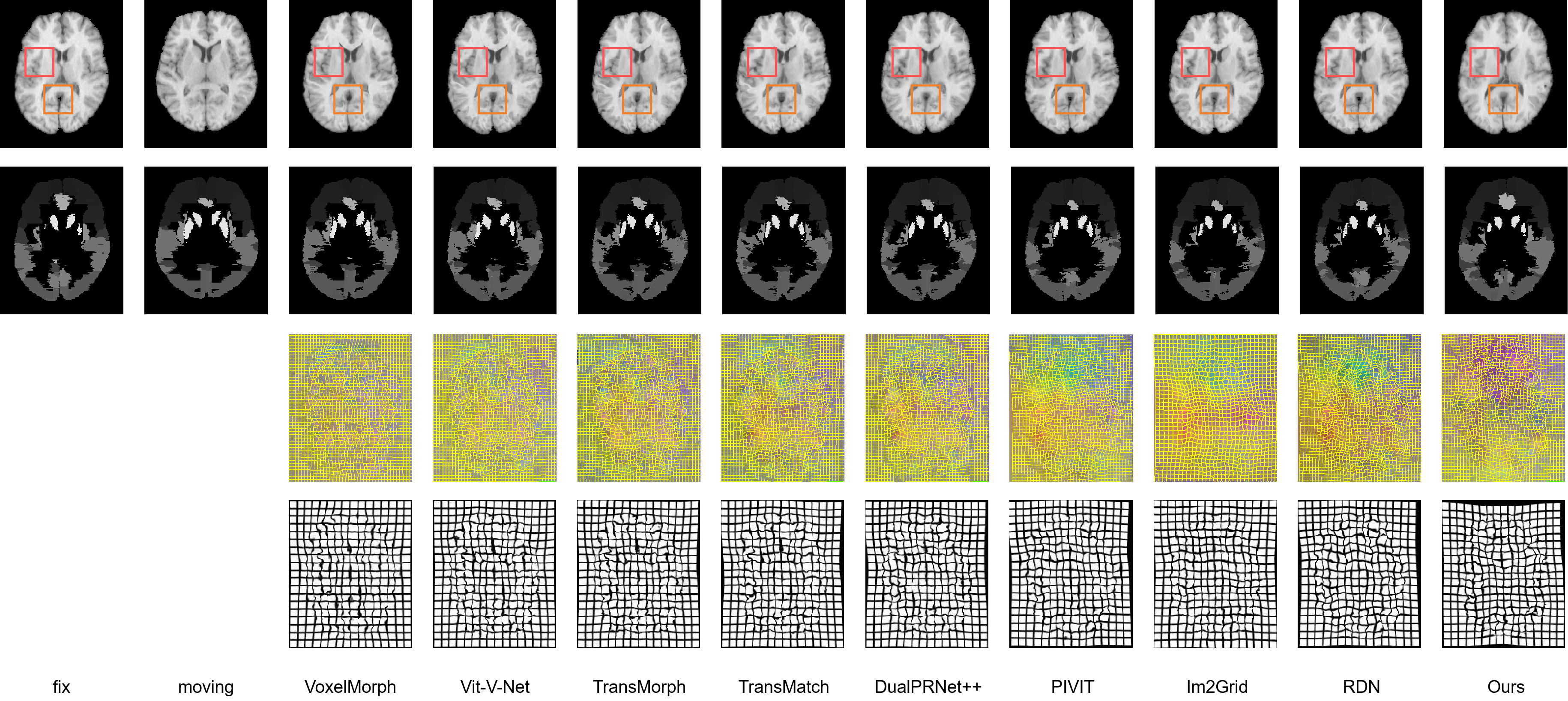}}
\caption{The visualization of the registration results from different methods on LPBA40 datasets. The two leftmost columns represent the original image and segmentation of the fixed and moving images, respectively. From top to bottom, the figure displays the warped images, warped segmentations, registration fields in grid format, and the deformed grids for different methods.}
\label{fig-model}
\end{center}
\end{figure*}

\subsection{Loss Function}
In this section, we introduce the training loss function of the unsupervised registration network FF-PNet. It uses normalized cross correlation (NCC) for training, evaluating similarity between the fixed image \(I_f\) and the warped moving image \(I_w = I_m \circ \phi\): 
\begin{equation}
\begin{split}
&\mathcal{L}_{sim}(I_f, I_{w}) = \\
&- \sum_{p \in \Omega} \frac{\sum_{p_i} \left[ I_f(p_i) - \overline{I_f}(p) \right] \left[ I_{w}(p_i) - \overline{I_{w}}(p) \right]}{\sqrt{\sum_{p_i} \left[ I_f(p_i) - \overline{I_f}(p) \right]^2 \sum_{p_i} \left[ I_{w}(p_i) - \overline{I_{w}}(p) \right]^2}}
\end{split}
\end{equation}

The notation \(\Omega\) represents the entire volumetric domain. For a voxel \(p\), \(p_i\) indicates the local neighborhood within a \(n^3\) (with \(n = 9\) in our scenario) volumetric patch at this voxel. Additionally, \(\bar{I_f}(p)\) and \(\bar{I_w}(p)\) denote the average value in their respective local neighborhoods. For the \(\mathcal{L}_{reg}\), we impose a diffusion regularizer on the \(\phi\) to encourage its smoothness:
\begin{equation}
\begin{split}
    \mathcal{L}_{reg}(\phi) = \sum_{p \in \Omega} ||\nabla \phi(p)||^2
\end{split}
\end{equation}

Therefore, the total loss is:
\begin{equation}\label{total}
\begin{split}
    \mathcal{L}_{total} = \mathcal{L}_{sim}(I_f, I_{m \circ \phi}) + \lambda \mathcal{L}_{reg}(\phi)
\end{split}
\end{equation}

where \(\lambda\) is the hyperparameters used to balance the contribution of loss functions.

\section{experiments}
\subsection{Datasets and Evaluation Metrics}
We evaluated the effectiveness of the proposed network on two publicly available brain MRI datasets, including LPBA and OASIS. The LPBA dataset contains 40 T1-weighted MRI volumes, each with 54 manually labeled regions of interest (ROI). The MRI volumes in LPBA were strictly pre-aligned with MNI 305. After centering, the volumes were cropped to 160×192×160 (1mm×1mm×1mm). Thirty volumes (30×29 pairs) were used for training, and ten volumes (10×9 pairs) were used for testing. The OASIS dataset contains 451 T1-weighted MRI volumes, and the image size is uniformly 160×192×224. Among them, 394 MRI images were used for training, and 19 and 38 MRI images were used for validation and testing, respectively.


In this study, the Dice similarity coefficient (DSC), 95th percentile of the maximum Hausdorff distance (HD95), and logarithmic standard deviation of the Jacobian determinant (SDlogJ) of the displacement field were employed to quantitatively evaluate the proposed model. Specifically, HD95 serves to measure the similarity in regional contours and assess the quality of the predicted non-rigid deformation field by quantifying the percentage of non-rigid voxels. Optimal registration outcomes are characterized by higher DSC values alongside lower HD95 and SDlogJ values, thereby enabling a comprehensive assessment of the model's performance in image registration tasks.

\subsection{Experimental Settings}
The deep neural network constructed in this study is implemented based on the PyTorch framework and uses the Adam algorithm for network parameter optimization. During the training process, the learning rate is set to \(10^{-4}\), adopts a batch size of 1, and performs 500 rounds of iterations to ensure model convergence. All experimental training and testing work is completed on an NVIDIA Tesla V100 GPU equipped with 32 GB of memory. \(\lambda\) in \eqref{total} is set to 1.

\subsection{Experimental Results and Analysis}
Table \ref{tab:lpba} presents the comparative results of FF-PNet against other methods on the LPBA dataset. When compared with classical UNet-based methods such as VoxelMorph and TransMorph, FF-PNet improves DSC by 12.21\% and 6.76\% respectively, demonstrating stronger capability in preserving anatomical structural similarity. In contrast to pyramid-based approaches like DualPRNet++ and Im2Grid, the DSC metric of FF-PNet increases by 2.3 and 1.9 percentage points, highlighting the effectiveness of the multi-scale feature fusion module. For the OASIS dataset, as detailed in Table \ref{tab:oasis}, FF-PNet outperforms significant advantages over the highly performant method H-ViT in all three evaluation metrics, further validating the model’s consistent enhancement in generalization ability and registration accuracy across different datasets.
\begin{table}[htbp]
    \centering
    \caption{The numerical results of different registration methods on the LPBA(56 rois) dataset.}
    \begin{tabular}{cccc}
        \toprule
        Methods & DSC $\uparrow$ & SDlogJ $\downarrow$ & HD95(mm) $\downarrow$ \\
        \hline
        Affine only & 0.548$\pm$0.041 & - & - \\
        VoxelMorph \cite{balakrishnan2019voxelmorph} & 0.647$\pm$0.030 & 0.364$\pm$0.020 & 4.881$\pm$0.596 \\
        ViT-V-Net \cite{chen2021vit} & 0.658$\pm$0.027 & 0.394$\pm$0.025 & 4.677$\pm$0.485 \\
        TransMorph \cite{chen2022transmorph} & 0.680$\pm$0.024 & 0.370$\pm$0.024 & 4.507$\pm$0.447 \\
        TransMatch \cite{chen2023transmatch} & 0.678$\pm$0.025 & 0.347$\pm$0.014 & 4.460$\pm$0.458 \\
        DualPRNet++ \cite{kang2022dual}& 0.703$\pm$0.017 & 0.275$\pm$0.011 & 4.097$\pm$0.352 \\
        PIVIT \cite{ma2023pivit} & 0.706$\pm$0.010 & \textbf{0.246$\pm$0.007} & 3.932$\pm$0.256 \\
        Im2Grid \cite{liu2022coordinate} & 0.707$\pm$0.011 & 0.270$\pm$0.012 & 3.952$\pm$0.290 \\
        RDN \cite{hu2022recursive} & 0.720$\pm$0.011 & 0.336$\pm$0.049 & \textbf{3.756$\pm$0.294} \\
        \hline
        FF-PNet (\textbf{Ours}) & \textbf{0.726$\pm$0.010} & 0.297$\pm$0.010 & 3.767$\pm$0.282 \\
        \toprule
    \end{tabular}
\label{tab:lpba}
\end{table}

\begin{table}[htbp]
    \centering
    \caption{The numerical results of different registration methods on the OASIS(35 rois) dataset.}
    \begin{tabular}{cccc}
        \toprule
        Methods & DSC $\uparrow$ & SDlogJ $\downarrow$ & HD95(mm) $\downarrow$\\
        \midrule
        VoxelMorph \cite{balakrishnan2019voxelmorph} &  0.844$\pm$0.013 & 0.509$\pm$0.021 & 1.221$\pm$0.144 \\
        ViT-V-Net \cite{chen2021vit} &  0.854$\pm$0.015 & 0.513$\pm$0.023 & 1.178$\pm$0.137 \\
        TransMorph \cite{chen2022transmorph} &  0.855$\pm$0.014 & 0.469$\pm$0.023 & 1.131$\pm$0.122 \\
        TransMorph-Large \cite{chen2022transmorph} &  0.859$\pm$0.014 & 0.465$\pm$0.019 & 1.125$\pm$0.135 \\
        PIVIT \cite{ma2023pivit} & 0.840$\pm$0.015 & \textbf{0.414$\pm$0.022} & 1.244$\pm$0.146 \\
        DualPRNet++ \cite{kang2022dual} &  0.879$\pm$0.015 & 0.443$\pm$0.025 & 1.090$\pm$0.146 \\
        H-ViT \cite{ghahremani2024h}&  0.876$\pm$0.014 & 0.539$\pm$0.069 & 1.301$\pm$0.264 \\
        \hline
        FF-PNet (\textbf{Ours}) & \textbf{0.884$\pm$0.015} & 0.466$\pm$0.023 & \textbf{1.059$\pm$0.162} \\
        \toprule
    \end{tabular}
\label{tab:oasis}
\end{table}

\begin{figure}[htbp]
\centering
\includegraphics[width=1\linewidth]{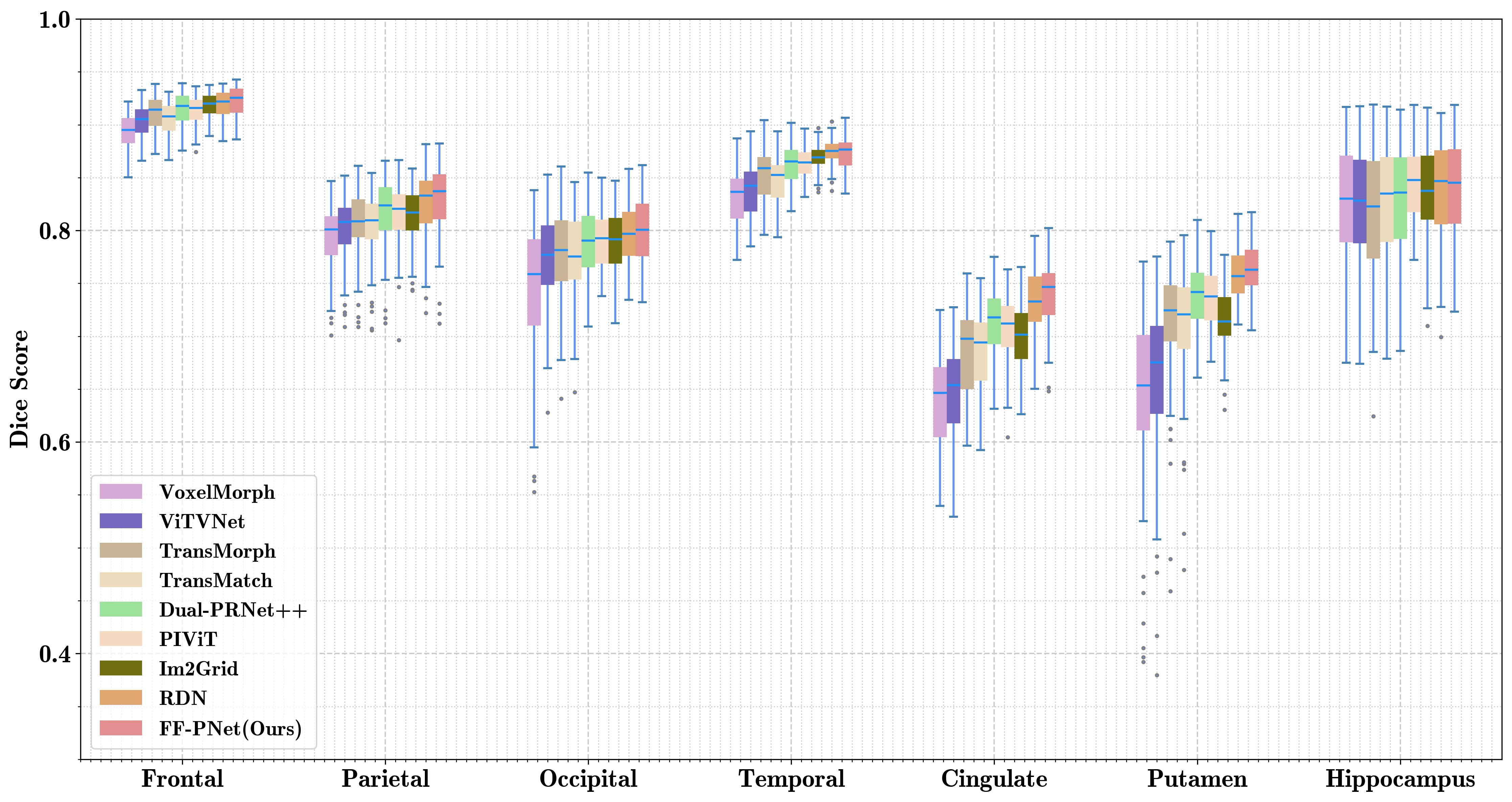}
\caption{Box plots illustrate the distribution of DSC across seven brain regions—frontal, parietal, occipital, temporal, cingulate, hippocampus, and putamen—generated by different registration methods on the LPBA dataset.}
\end{figure}
\begin{figure}[htbp]
\centering
\includegraphics[width=1\linewidth]{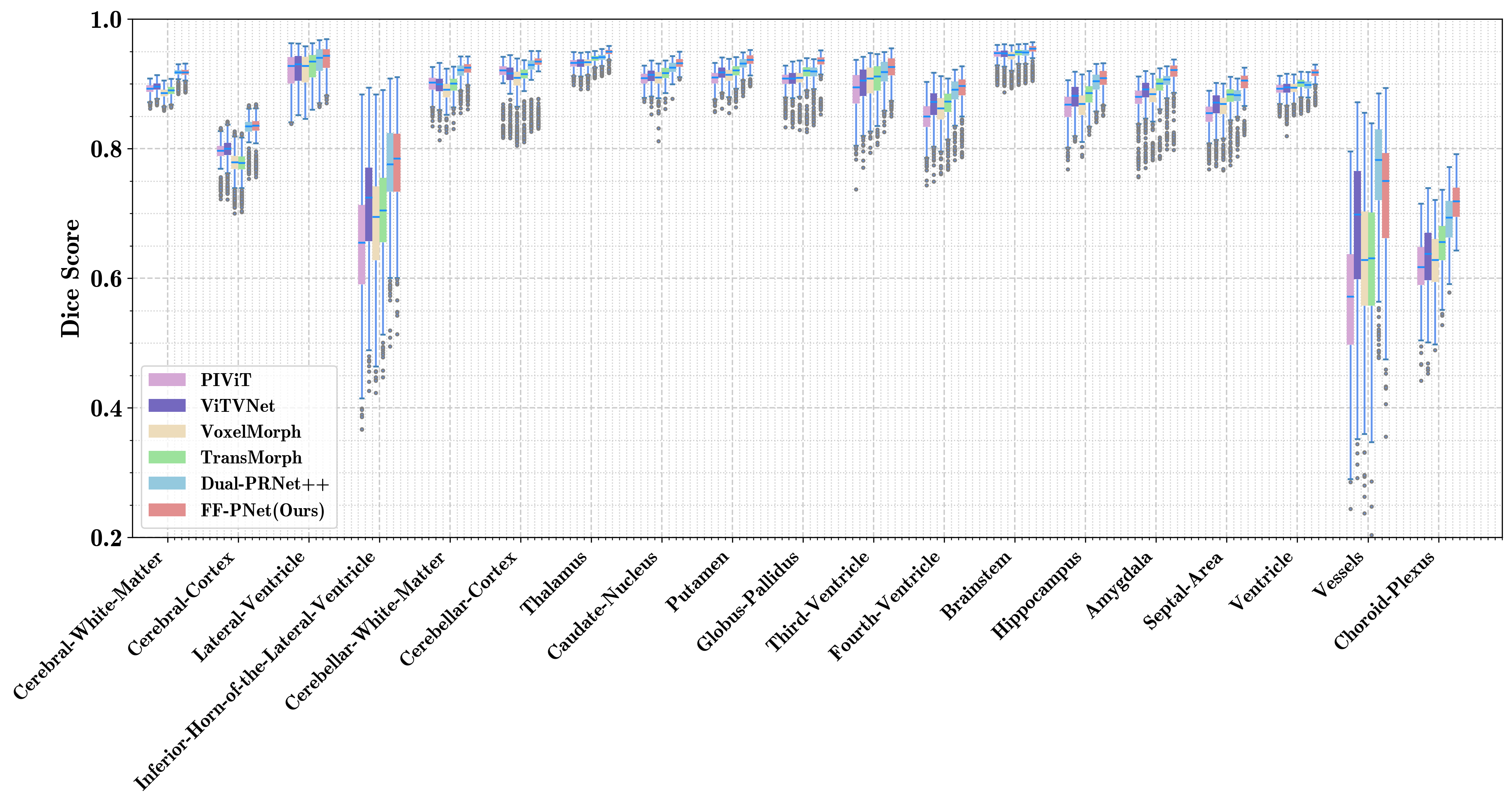}
\caption{Box plots showing Dice scores for different brain MR substructures using the proposed FF-PNet and existing image registration methods.}
\end{figure}

\subsection{Ablation experiments}
In this subsection, we ablate each component in our FF-PNet to show its impact. All experiments were conducted on the LPBA40 dataset with identical hyperparameters to ensure fair and consistent comparisons.

\textbf{Ablation Study on Modules.}
To investigate the impacts of the Residual Feature Fusion Module (RFFM) and Residual Deformation Field Fusion Module (RDFFM) on model performance, as well as the synergistic effects when the two modules are combined, we conducted ablation experiments on a three-layer pyramid network architecture. As demonstrated in Table \ref{table-fusion}, the individual adoption of either RFFM or RDFFM substantially improved image registration accuracy, validating the necessity of these residual fusion modules in model design. When the two modules synergistically operate, our proposed full framework FF-PNet achieves superior performance across all evaluation metrics, exemplifying the complementary advantages of the dual-stream fusion mechanism.
\begin{table}[htbp]
  \setlength{\abovecaptionskip}{0.8pt}
  \setlength{\belowcaptionskip}{5pt} 
  \centering
  \caption{Results of ablation experiments for fusion module in FF-PNet.}
  \begin{tabular}{c c c c c} 
    \toprule
    RFFM & RDFFM & DSC $\uparrow$ & HD95(mm) $\downarrow$ & Params(M)\\ 
    \midrule
    $\times$ & $\times$ & 0.669$\pm$0.026 & 4.592$\pm$0.481 & 6.770 \\ 
    $\times$ & \checkmark & 0.704$\pm$0.014 & 4.012$\pm$0.302 & 6.820 \\
    \checkmark & $\times$ & 0.717$\pm$0.015 & 3.888$\pm$0.334 & 8.920 \\
    \checkmark & \checkmark & \textbf{0.726$\pm$0.010} & \textbf{3.767$\pm$0.282} & 8.920 \\
    \bottomrule
  \end{tabular}
\label{table-fusion}
\end{table}

\textbf{Ablation Study on Layers.}
To validate the three-layer architecture of FF-PNet, ablation studies were conducted on a four-layer network integrated with RFFM and RDFFM. Table \ref{table-layer} shows the four-layer pure convolutional pyramid outperformed the three-layer structure without fusion modules. However, with the dual-stream fusion module applied, the three-layer architecture significantly improved registration accuracy by enhancing mid-level feature interaction to capture critical local spatial correspondences—even with limited high-level semantic information. This highlights the optimal synergy between the three-layer design and fusion module in balancing semantic and spatial information.
\begin{table}[htbp]
  \setlength{\abovecaptionskip}{0.8pt}
  \setlength{\belowcaptionskip}{5pt} 
  \setlength{\tabcolsep}{5pt}
  \centering
  \caption{Results of ablation experiments for different layers on FF-PNet.}
  \begin{tabular}{c c c c c c} 
    \toprule
    Layers & RFFM & RDFFM & DSC $\uparrow$ & HD95(mm) $\downarrow$ & Params(M)\\ 
    \midrule
    3 & $\times$ & $\times$ & 0.669$\pm$0.026 & 4.592$\pm$0.481 & 6.770 \\ 
    3 & \checkmark & \checkmark & \textbf{0.726$\pm$0.010} & \textbf{3.767$\pm$0.282} & 8.920 \\
    4 & $\times$ & $\times$ & 0.667$\pm$0.027 & 4.622$\pm$0.494 & 27.67 \\
    4 & \checkmark & \checkmark & 0.726$\pm$0.011 & 3.791$\pm$0.279 & 36.78 \\
    \bottomrule
  \end{tabular}
\label{table-layer}
\end{table}

\textbf{Ablation Study on Channels.}
Ablation studies on the initial channel numbers of FF-PNet were conducted, investigating four configurations: 16, 32, 48, and 64. Experimental findings in Table \ref{table-dim} demonstrate that increasing channel numbers leads to a substantial rise in parametric complexity, whereas moderate reductions in channel counts do not result in significant declines in registration accuracy—validating the efficient feature decoding capability of the dual-stream fusion module. Notably, when channel numbers exceed 48, parametric complexity surges without a corresponding notable improvement in DSC. Balancing model complexity and performance, 48 is selected as the optimal initial channel number for FF-PNet, ensuring adequate feature representation while avoiding unnecessary computational overhead.
\begin{table}[htbp]
    \centering
    \caption{Results of selecting different channels on FF-PNet.}
    \setlength{\tabcolsep}{10pt}
    \begin{tabular}{cccc}
        \toprule 
        Channel & DSC $\uparrow$ & SDlogJ $\downarrow$ & Params(M) \\
        \midrule 
        16 & 0.707$\pm$0.015 & 0.300$\pm$0.009 & 0.933 \\
        32 & 0.720$\pm$0.012 & \textbf{0.285$\pm$0.011} & 3.966 \\
        48 & \textbf{0.726$\pm$0.010} & 0.297$\pm$0.010 & 8.920 \\
        64 & 0.726$\pm$0.012 & 0.312$\pm$0.012 & 15.85 \\
        \bottomrule 
    \end{tabular}
    \label{table-dim}
\end{table}

\section{conclusion}
This paper presents FF-PNet, a network specifically designed for medical image registration tasks, featuring a three-layer pyramid architecture and employing a dual-stream fusion module for feature decoding. Aiming at the distinct characteristics of features and deformation fields, FF-PNet innovatively introduces the Residual Feature Fusion Module (RFFM) and the Residual Deformation Field Fusion Module (RDFFM). These two modules extract differential residual flows and perform efficient fusion, effectively addressing the challenge of large-deformation image registration without relying on attention mechanisms or transformers. Through the parallel collaboration of RFFM and RDFFM, the network organically integrates the detail-capturing capabilities of the UNet architecture with the advantages of multi-scale contextual modeling offered by the pyramid architecture, constructing a coarse-to-fine hierarchical registration framework. Notably, the proposed fusion modules exhibit strong generality and can be flexibly embedded into various network architectures to enhance image registration performance. Experimental results on multiple benchmark datasets demonstrate that FF-PNet significantly outperforms prevalent methods, showcasing superior registration accuracy and robust generalization capability.

\vspace{12pt}

\bibliographystyle{IEEEtran}
\small\bibliography{reference}

\begin{thebibliography}{10}
\providecommand{\url}[1]{#1}
\csname url@samestyle\endcsname
\providecommand{\newblock}{\relax}
\providecommand{\bibinfo}[2]{#2}
\providecommand{\BIBentrySTDinterwordspacing}{\spaceskip=0pt\relax}
\providecommand{\BIBentryALTinterwordstretchfactor}{4}
\providecommand{\BIBentryALTinterwordspacing}{\spaceskip=\fontdimen2\font plus
\BIBentryALTinterwordstretchfactor\fontdimen3\font minus \fontdimen4\font\relax}
\providecommand{\BIBforeignlanguage}[2]{{%
\expandafter\ifx\csname l@#1\endcsname\relax
\typeout{** WARNING: IEEEtran.bst: No hyphenation pattern has been}%
\typeout{** loaded for the language `#1'. Using the pattern for}%
\typeout{** the default language instead.}%
\else
\language=\csname l@#1\endcsname
\fi
#2}}
\providecommand{\BIBdecl}{\relax}
\BIBdecl

\bibitem{matinfar2023sonification}
S.~Matinfar, M.~Salehi, D.~Suter, M.~Seibold, S.~Dehghani, N.~Navab, F.~Wanivenhaus, P.~F{\"u}rnstahl, M.~Farshad, and N.~Navab, ``Sonification as a reliable alternative to conventional visual surgical navigation,'' \emph{Scientific Reports}, vol.~13, no.~1, p. 5930, 2023.

\bibitem{sotiras2013deformable}
A.~Sotiras, C.~Davatzikos, and N.~Paragios, ``Deformable medical image registration: A survey,'' \emph{IEEE transactions on medical imaging}, vol.~32, no.~7, pp. 1153--1190, 2013.

\bibitem{siddiqui2015comparison}
M.~M. Siddiqui, S.~Rais-Bahrami, B.~Turkbey, A.~K. George, J.~Rothwax, N.~Shakir, C.~Okoro, D.~Raskolnikov, H.~L. Parnes, W.~M. Linehan \emph{et~al.}, ``Comparison of mr/ultrasound fusion--guided biopsy with ultrasound-guided biopsy for the diagnosis of prostate cancer,'' \emph{Jama}, vol. 313, no.~4, pp. 390--397, 2015.

\bibitem{klein2009evaluation}
A.~Klein, J.~Andersson, B.~A. Ardekani, J.~Ashburner, B.~Avants, M.-C. Chiang, G.~E. Christensen, D.~L. Collins, J.~Gee, P.~Hellier \emph{et~al.}, ``Evaluation of 14 nonlinear deformation algorithms applied to human brain mri registration,'' \emph{Neuroimage}, vol.~46, no.~3, pp. 786--802, 2009.

\bibitem{rueckert2006diffeomorphic}
D.~Rueckert, P.~Aljabar, R.~A. Heckemann, J.~V. Hajnal, and A.~Hammers, ``Diffeomorphic registration using b-splines,'' in \emph{Medical Image Computing and Computer-Assisted Intervention--MICCAI 2006: 9th International Conference, Copenhagen, Denmark, October 1-6, 2006. Proceedings, Part II 9}.\hskip 1em plus 0.5em minus 0.4em\relax Springer, 2006, pp. 702--709.

\bibitem{fu2020deep}
Y.~Fu, Y.~Lei, T.~Wang, W.~J. Curran, T.~Liu, and X.~Yang, ``Deep learning in medical image registration: a review,'' \emph{Physics in Medicine \& Biology}, vol.~65, no.~20, p. 20TR01, 2020.

\bibitem{balakrishnan2018unsupervised}
G.~Balakrishnan, A.~Zhao, M.~R. Sabuncu, J.~Guttag, and A.~V. Dalca, ``An unsupervised learning model for deformable medical image registration,'' in \emph{Proceedings of the IEEE conference on computer vision and pattern recognition}, 2018, pp. 9252--9260.

\bibitem{ronneberger2015u}
O.~Ronneberger, P.~Fischer, and T.~Brox, ``U-net: Convolutional networks for biomedical image segmentation,'' in \emph{Medical image computing and computer-assisted intervention--MICCAI 2015: 18th international conference, Munich, Germany, October 5-9, 2015, proceedings, part III 18}.\hskip 1em plus 0.5em minus 0.4em\relax Springer, 2015, pp. 234--241.

\bibitem{hoopes2021hypermorph}
A.~Hoopes, M.~Hoffmann, B.~Fischl, J.~Guttag, and A.~V. Dalca, ``Hypermorph: Amortized hyperparameter learning for image registration,'' in \emph{Information Processing in Medical Imaging: 27th International Conference, IPMI 2021, Virtual Event, June 28--June 30, 2021, Proceedings 27}.\hskip 1em plus 0.5em minus 0.4em\relax Springer, 2021, pp. 3--17.

\bibitem{zhang2024dual}
Y.~Zhang, S.~Guo, D.~Shi, and J.~Xiang, ``Dual decoder unet with contrastive learning for brain image registration,'' in \emph{2024 IEEE International Conference on Bioinformatics and Biomedicine (BIBM)}.\hskip 1em plus 0.5em minus 0.4em\relax IEEE, 2024, pp. 2922--2927.

\bibitem{zhang2023ccac}
Y.~Zhang, D.~Shi, J.~Wang, J.~Xiang, and W.~Zhang, ``Ccac: Contrastive learning with channel attention and contour loss for brain image registration,'' in \emph{2023 IEEE International Conference on Bioinformatics and Biomedicine (BIBM)}.\hskip 1em plus 0.5em minus 0.4em\relax IEEE, 2023, pp. 1704--1709.

\bibitem{balakrishnan2019voxelmorph}
G.~Balakrishnan, A.~Zhao, M.~R. Sabuncu, J.~Guttag, and A.~V. Dalca, ``Voxelmorph: a learning framework for deformable medical image registration,'' \emph{IEEE transactions on medical imaging}, vol.~38, no.~8, pp. 1788--1800, 2019.

\bibitem{dosovitskiy2020image}
A.~Dosovitskiy, L.~Beyer, A.~Kolesnikov, D.~Weissenborn, X.~Zhai, T.~Unterthiner, M.~Dehghani, M.~Minderer, G.~Heigold, S.~Gelly \emph{et~al.}, ``An image is worth 16x16 words: Transformers for image recognition at scale,'' \emph{arXiv preprint arXiv:2010.11929}, 2020.

\bibitem{chen2022transmorph}
J.~Chen, E.~C. Frey, Y.~He, W.~P. Segars, Y.~Li, and Y.~Du, ``Transmorph: Transformer for unsupervised medical image registration,'' \emph{Medical image analysis}, vol.~82, p. 102615, 2022.

\bibitem{vaswani2017attention}
A.~Vaswani, N.~Shazeer, N.~Parmar, J.~Uszkoreit, L.~Jones, A.~N. Gomez, {\L}.~Kaiser, and I.~Polosukhin, ``Attention is all you need,'' \emph{Advances in neural information processing systems}, vol.~30, 2017.

\bibitem{kang2022dual}
M.~Kang, X.~Hu, W.~Huang, M.~R. Scott, and M.~Reyes, ``Dual-stream pyramid registration network,'' \emph{Medical image analysis}, vol.~78, p. 102379, 2022.

\bibitem{liu2022coordinate}
Y.~Liu, L.~Zuo, S.~Han, Y.~Xue, J.~L. Prince, and A.~Carass, ``Coordinate translator for learning deformable medical image registration,'' in \emph{International workshop on multiscale multimodal medical imaging}.\hskip 1em plus 0.5em minus 0.4em\relax Springer, 2022, pp. 98--109.

\bibitem{wang2023modet}
H.~Wang, D.~Ni, and Y.~Wang, ``Modet: learning deformable image registration via motion decomposition transformer,'' in \emph{International Conference on Medical Image Computing and Computer-Assisted Intervention}.\hskip 1em plus 0.5em minus 0.4em\relax Springer, 2023, pp. 740--749.

\bibitem{ma2023pivit}
T.~Ma, X.~Dai, S.~Zhang, and Y.~Wen, ``Pivit: Large deformation image registration with pyramid-iterative vision transformer,'' in \emph{International Conference on Medical Image Computing and Computer-Assisted Intervention}.\hskip 1em plus 0.5em minus 0.4em\relax Springer, 2023, pp. 602--612.

\bibitem{dalca2018andm}
A.~Dalca, G.~Balakrishnan, and J.~Guttag, ``andm. r. sabuncu,“unsupervised learning for fast probabilistic diffeomorphic registration,”,'' in \emph{Proc. Int. Conf. Med. Image Comput. Comput.-Assist. Intervention}, 2018, pp. 729--738.

\bibitem{liu2021swin}
Z.~Liu, Y.~Lin, Y.~Cao, H.~Hu, Y.~Wei, Z.~Zhang, S.~Lin, and B.~Guo, ``Swin transformer: Hierarchical vision transformer using shifted windows,'' in \emph{Proceedings of the IEEE/CVF international conference on computer vision}, 2021, pp. 10\,012--10\,022.

\bibitem{zhu2022swin}
Y.~Zhu and S.~Lu, ``Swin-voxelmorph: A symmetric unsupervised learning model for deformable medical image registration using swin transformer,'' in \emph{International Conference on Medical Image Computing and Computer-Assisted Intervention}.\hskip 1em plus 0.5em minus 0.4em\relax Springer, 2022, pp. 78--87.

\bibitem{zhao2019recursive}
S.~Zhao, Y.~Dong, E.~I. Chang, Y.~Xu \emph{et~al.}, ``Recursive cascaded networks for unsupervised medical image registration,'' in \emph{Proceedings of the IEEE/CVF international conference on computer vision}, 2019, pp. 10\,600--10\,610.

\bibitem{lv2022joint}
J.~Lv, Z.~Wang, H.~Shi, H.~Zhang, S.~Wang, Y.~Wang, and Q.~Li, ``Joint progressive and coarse-to-fine registration of brain mri via deformation field integration and non-rigid feature fusion,'' \emph{IEEE Transactions on Medical Imaging}, vol.~41, no.~10, pp. 2788--2802, 2022.

\bibitem{chen2021vit}
J.~Chen, Y.~He, E.~C. Frey, Y.~Li, and Y.~Du, ``Vit-v-net: Vision transformer for unsupervised volumetric medical image registration,'' \emph{arXiv preprint arXiv:2104.06468}, 2021.

\bibitem{chen2023transmatch}
Z.~Chen, Y.~Zheng, and J.~C. Gee, ``Transmatch: a transformer-based multilevel dual-stream feature matching network for unsupervised deformable image registration,'' \emph{IEEE transactions on medical imaging}, vol.~43, no.~1, pp. 15--27, 2023.

\bibitem{hu2022recursive}
B.~Hu, S.~Zhou, Z.~Xiong, and F.~Wu, ``Recursive decomposition network for deformable image registration,'' \emph{IEEE Journal of Biomedical and Health Informatics}, vol.~26, no.~10, pp. 5130--5141, 2022.

\bibitem{ghahremani2024h}
M.~Ghahremani, M.~Khateri, B.~Jian, B.~Wiestler, E.~Adeli, and C.~Wachinger, ``H-vit: A hierarchical vision transformer for deformable image registration,'' in \emph{Proceedings of the IEEE/CVF Conference on Computer Vision and Pattern Recognition}, 2024, pp. 11\,513--11\,523.

\end{thebibliography}

\end{document}